\documentclass[letterpaper, 10 pt, conference]{ieeeconf}  % Comment this line out if you need a4paper

\usepackage[style=ieee,backend=biber]{biblatex}
\usepackage{graphicx}
\usepackage[ruled, linesnumbered]{algorithm2e}
\usepackage[utf8]{inputenc}

\makeatletter
\let\MYcaption\@makecaption
\makeatother

\usepackage[font=footnotesize]{subcaption}

\makeatletter
\let\@makecaption\MYcaption
\makeatother

\usepackage{amsmath}

\IEEEoverridecommandlockouts                              % This command is only needed if 
\SetKwRepeat{Do}{do}{while}

\title{\LARGE \bf
Ultrasound-Guided Robotic Navigation \\ with Deep Reinforcement Learning}
\author{Hannes Hase$^{*,1}$, Mohammad Farid Azampour$^{*,1,2}$, Maria Tirindelli$^{1}$,\\ Magdalini Paschali$^{1}$, Walter Simson$^{1}$, Emad Fatemizadeh$^{2}$ and Nassir Navab$^{1,3}$% <-this % stops a space
\thanks{$^{*}$These authors contributed equally to this work}%
\thanks{$^{1}$Computer Aided Medical Procedures, Technische Universit\"at M\"unchen, Munich, Germany
        {\tt\small hannes.hase@tum.de}}%
\thanks{$^{2}$Sharif University of Technology, Tehran, Iran
        {\tt\small mf.azampour@tum.de}}%
\thanks{$^{3}$Computer Aided Medical Procedures, John Hopkins University, Baltimore, MD, USA}
}

\begin{document}

% Uncomment for removing borders of figure boxes!
\setlength{\fboxrule}{0pt}

\maketitle
\thispagestyle{empty}
\pagestyle{empty}

%%%%%%%%%%%%%%%%%%%%%%%%%%%%%%%%%%%%%%%%%%%%%%%%%%%%%%%%%%%%%%%%%%%%%%%%%%%%%%%%
\begin{abstract}

In this paper we introduce the first reinforcement learning (RL) based robotic navigation method which utilizes ultrasound (US) images as an input. Our approach combines state-of-the-art RL techniques, specifically deep Q-networks (DQN) with memory buffers and a binary classifier for deciding when to terminate the task. 

Our method is trained and evaluated on an in-house collected data-set of 34 volunteers and when compared to pure RL and supervised learning (SL) techniques, it performs substantially better, which highlights the suitability of RL navigation for US-guided procedures. When testing our proposed model, we obtained a 82.91\% chance of navigating correctly to the sacrum
from 165 different starting positions on 5 different unseen simulated environments.

\end{abstract}

%%%%%%%%%%%%%%%%%%%%%%%%%%%%%%%%%%%%%%%%%%%%%%%%%%%%%%%%%%%%%%%%%%%%%%%%%%%%%%%%
\section{Introduction}

The rise of robotics and their gradual permeation into the field of medicine is a revolution on its own. By integrating robotic systems in the medical work-space, doctors are enabled to treat individual patients in a more efficient, safer and less morbid way. However, end-to-end automated approaches are constrained by the adaptability to unexpected situations and the poor judgment of robotic systems~\cite{taylor2006perspective}.

With ever-improving ultrasound (US) technology, US is being increasingly used in diagnostics and interventions. Unlike other modalities like computed tomography (CT), US provides real-time dynamic physiologic information while being radiation free and comparatively cheap. Yet, the quality of an US image suffers from artifacts such as speckle and clutter, has a low signal to noise ratio and is strongly subject dependent~\cite{hindi2013artifacts}. Another downside is the high inter-observer variability when acquiring US images, which calls for trained sonographers to guarantee clinically relevant images. It is the lack of specialists that opens the need for robotic imaging techniques~\cite{jintao2019robotic}. The mentioned difficulties associated with US imaging make the task of autonomous US navigation extremely challenging.

Robotic ultrasound (rUS) in the medical field has been investigated to improve working conditions for doctors and also to increase the accuracy of interventions~\cite{esteban2018robotic, hennersperger2016towards}.   
Tirindelli et al. in ~\cite{tirindelli2020forceultrasound} attempt to automate spinal navigation by using a combination of force data and US image. However, this procedure still requires to be set-up by a technician. Automatic navigation towards specific positions without any human intervention on the human body is still not resolved, to the best of our knowledge.

Reinforcement learning offers an interesting and novel approach, as it excels at sequential decision making and exploratory tasks~\cite{Sutton1998}.
Reinforcement learning has shown superhuman performance on Atari games~\cite{mnih2013playing} in which the agent only decides what to do based on visual input. This has already been translated to real-life applications in visual robotic manipulation, such as the general task of grasping~\cite{quillen2018deep} or in visual navigation for humanoid robots playing soccer~\cite{lobos2018visual}.
Even in the medical field, initial attempts have been made to exploit the strengths of RL. 
For instance,~\cite{alansary2019evaluating} proposes to use RL to find landmarks in fetal magnetic resonance imaging (MRI) scans, in order to improve 3D-imaging.

With the goal of expanding the applications of RL in the medical sector, we work towards the full automation of spinal navigation solely relying on US images for the decision making. 
Towards this end we propose a method using a combination of RL and supervised learning (SL) overcoming disadvantages of both approaches.

In detail, our contributions are:
\begin{enumerate}
    \item The acquisition of an in-house data-set of lower back US sweeps on volunteers using a robot for accurate tracking of the frames. 
    \item Training an RL agent on simulated lower-back environments to find correct views of the sacrum while navigating the environments only relying on US frames.
    % \item Reporting the results of a feasibility study on the application of RL for US based autonomous navigation in medical settings.
\end{enumerate}

\begin{figure*}[th]
  \centering
  \framebox{
  \centering
  \includegraphics[scale=0.46]{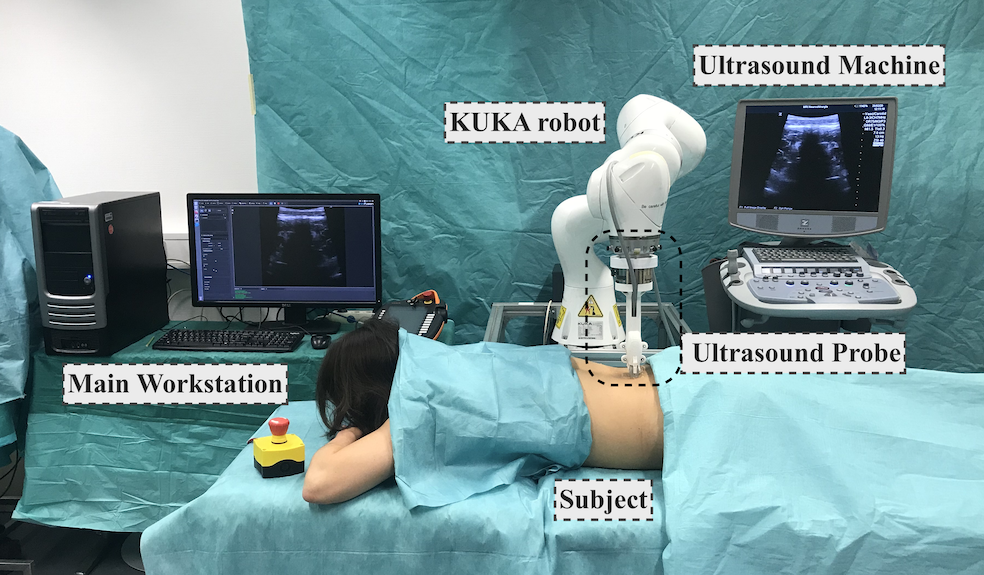}
  }
  \caption{Setup for robotic ultrasound acquisition. Ultrasound probe is attached to the robot end-effector using a 3D printed holder. Main workstation will store the frames acquired by the US machine alongside the tracking data from the robot.}
  \label{project_setup}
\end{figure*}

\section{Related Work}

\subsection{Deep Reinforcement Learning}

RL is one of the three main paradigms, of machine learning, alongside supervised and unsupervised learning~\cite{Sutton1998}.
In RL, an agent interacts with an environment and aims at maximizing an accumulated reward that results from its actions.
Arulkumaran et al. provides a comprehensive overview of the developments of deep reinforcement learning (DRL)~\cite{arulkumaran2017brief}. 
In RL an agent is trained to complete a task via specialization in goal-directed learning. An environment is modeled in which the agent can explore and associate actions with rewards and thus, learn how to achieve the defined goal~\cite{Sutton1998}.
For matters of this study, we discuss DRL further in the methodology section.

\subsection{Reinforcement Learning for Robotic Manipulation}
Vision-based robotic manipulation with reinforcement learning is first investigated in~\cite{zhang2015towards}. 
Zhang et al. train an agent to autonomously steer a robot to reach a target using raw pixels as the sole input. 
While training and testing using simulated environments provide promising results, their approach fails when transferred to real-world applications. 
In~\cite{quillen2018deep}, the authors propose a benchmark for the general task of grasping using popular RL methods like deep Q-learning (DQL) and deep deterministic policy gradient (DDPG). 
Based on their results, DQL translates into more stable agents in case of small data-sets, whereas Monte Carlo methods provide better results on larger sets. 
They report a success-rate of 50\% on a relatively small data-set of 10k samples.

\subsection{Reinforcement Learning in Medicine}
Chu et al. combine online SL and RL for improving the efficiency of breast cancer diagnosis in clinics on multi-modal data~\cite{chu2016adaptive}.
The online SL assesses breast cancer risk based on the available patient data and examinations.
The doctor then decides if the confidence of the diagnostic was high enough.
If the confidence is not enough, the RL part of the framework recommends the next best measurements or exams that would improve the diagnostics' confidence.

Initial exploratory works have experimented with visual RL for medical applications. 
Milletari et al.~\cite{milletari2019straight} successfully propose DRL to perform action suggestion for sonographer guidance. 
In this seminal work a DRL agent successfully learns a policy to guide inexperienced medical personnel to obtain clinically relevant cardiac ultrasound images of the parasternal long-axis view. 
The authors simulate the RL environments by projecting a grid on subjects' chests and populating the grids' sectors or bins with in-vivo US-frames collected on a set of volunteers. 
At inference time, the user acts as the agent and is provided motion recommendations by the RL-policy; manually closing the loop of navigation. 
Building on this work, we close the agent-policy loop by adding a robotic actuator to manipulate the ultrasound probe based on the RL-policy. Additionally, we improve the DQN by adding memory to the model and using a binary classifier for stopping.

\begin{figure*}[thpb]
  \centering
  \framebox{
  \centering
  \includegraphics[scale=0.42]{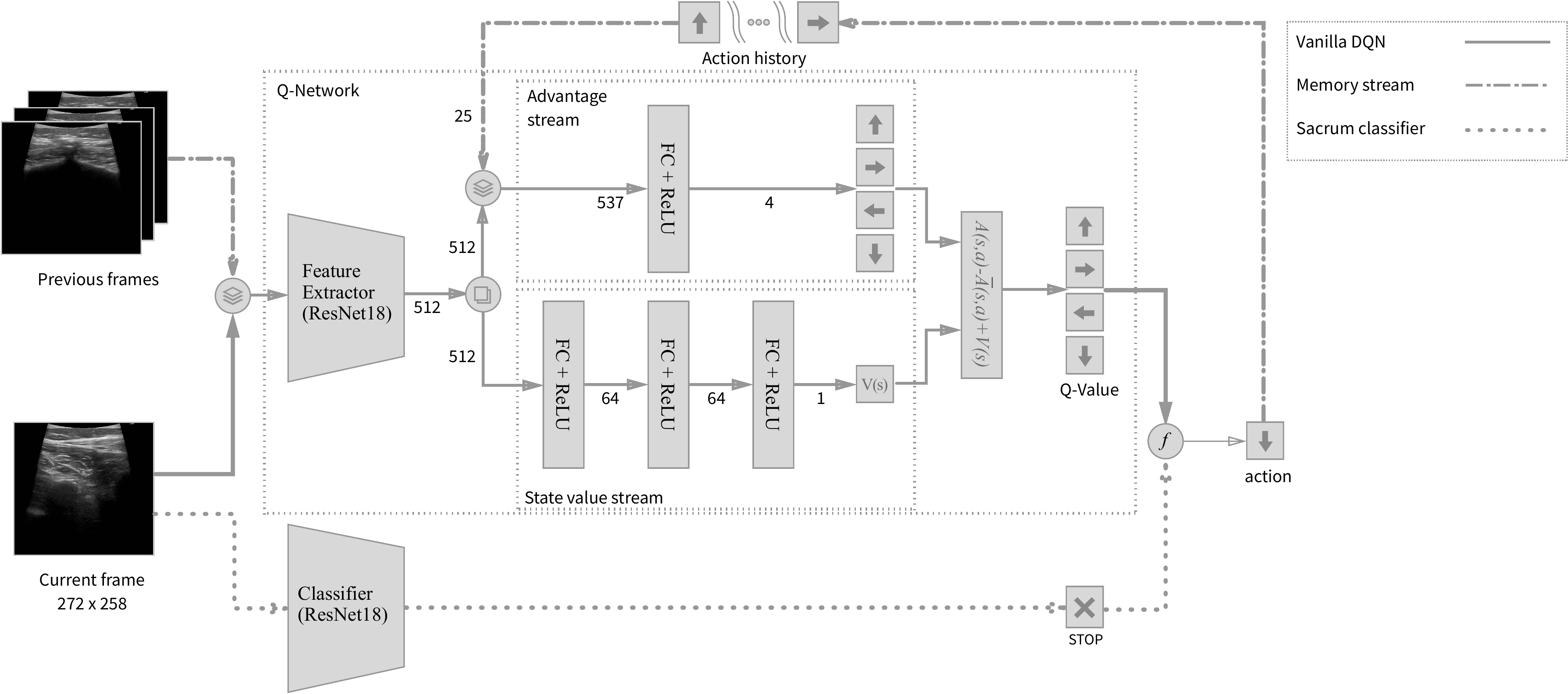}
  }
  \caption{Overall network architecture. The solid arrow represents the V-DQN. The broken and the dotted line, describe the changes introduced by M-DQN and MS-DQN in the V-DQN, respectively. When not using the binary classification network for stopping, the stop action becomes part of the Q-value layer as a fifth value.}
  \label{network_architecture}
\end{figure*}

\section{Methodology}

\subsection{Reinforcement Learning}

RL-problems are often modeled as Markov Decision Processes (MPD).
A MDP is a sequential decision problem for a fully observable, stochastic environment with a Markovian transition model and additive rewards.
It consists of a set of states $S$, a set of actions for each state $\mathcal{S}_a$, a transition model $P(s'|s,a)$ and a reward function $R(s)$~\cite{Sutton1998}.
In our work, the agent relies exclusively on visual input in the form of an US frame.
Thus, the agent does not explicitly know its state and needs to estimate it.
This turns the problem into a partially observable MDP (POMDP).

\subsection{Deep Q-Learning}

Q-Learning is a form of model-free off-policy RL that enables agents to learn optimal behavior in Markovian domains.
The agent learns to estimate Q-values, defined as the long term reward of performing a certain action in a given state~\cite{watkins1992q}.
An RL-agent is trained by exposing it to random transitions represented by the tuple $(s,a,r,s')$, where $s, s'$ are the states, $a$ is the chosen action and $r$ is the reward gained at step $t$ and $t+1$ respectively.
The transitions are acquired by the agent while interacting with the environment and stored in a replay memory to break temporal correlations.
The training batches are sampled from the replay memory and fed into the DQN for training.
The Q-values are learned by iteratively improving the estimates based on the results of the interaction with the environment, following the equation:
\begin{equation}
\label{eq:q-function}
Q(s,a) \leftarrow Q(s,a) + \alpha (r + \gamma \max_{a'} Q(s',a') - Q(s,a))
\end{equation}
where $\alpha$ corresponds to the learning rate and $\gamma$ to the discount factor.

When the model converges to an optimal solution, we get the optimal action for a state $s$ by doing $argmax(Q(s,a))$.

The main difficulty of Q-learning's traditional look-up table method is successfully learning in environments with large state-spaces.
Mnih et al.~\cite{mnih2013playing} propose Deep Q-learning (DQL) as  a solution to this issue by approximating the the Q-values with neural networks in the context of training a RL-agent to play Atari video-games.
% The performance of the introduced architecture substantially by introducing a series of modifications summarized in~\cite{hessel2018rainbow}.

We improved the base DQN by including:
\begin{enumerate}
    \item \textbf{Double Deep Q-Network (DDQN)}: The base DQN setup is difficult to train because the model's neural network (NN) is used for computing at the same time the prediction and the target, leading to the targets changing at each training step and making the training unstable. This is solved by copying the DQN into a second network referred to as the target network, where the weights are fixed and updated based on the current DQN's weights every N training steps. By doing this, we avoid Q-value over-estimations and achieve a more reliable training.~\cite{van2016deep}
    \item \textbf{Dueling DQN}: Wang et al. in~\cite{wang2015dueling} introduce the splitting of the Q-value estimation into two streams, as shown in Fig.~\ref{network_architecture}. One the one hand, the advantage-value stream $A(s,a)$, estimates the short-term reward that is achievable with each available action. On the other hand, the state-value stream estimates the long-term reward that is possible from that state. The Q-values are then computed as detailed in Eq.~\ref{eq: duelling-dqn}.
    \item \textbf{Prioritized Replay Memory}: The time-difference or TD-error is defined in Q-learning as: \begin{equation}
        TD = r + \gamma \max_{a'} Q_{target}(s',a') - Q(s,a)
    \end{equation} and represents a measure of how unsuspected the transition used for training is. When sampling transitions for training, the transitions probability of being selected is dependent on its TD-error. Hereby, transitions with relevant information are prioritized for training.~\cite{schaul2015prioritized}
\end{enumerate}{}

This setup we call \textit{V-DQN}. We define $I_t$ as the input frame at time $t$, $\phi(\cdot)$ as the feature extractor, $f_v$ and $f_A$ as the value and action advantage estimators, respectively. The Q-values of the V-DQN model are a function of the current frame following Eq.~\ref{eq: duelling-dqn}. 

\begin{align}\label{eq: duelling-dqn}
    V(s) &= f_v(\phi(I_t)) \nonumber\\
    A(s,a) &= f_A(\phi(I_t), a) \\
    Q(s,a) &= A(s,a)-\bar{A}(s,a)+V(s) \nonumber
\end{align}

In this work, we add two input streams of previous transitions in the environment. The first one corresponds to the previous frames, as done in~\cite{mnih2013playing}. For the second one, we adapt the method proposed by~\cite{yun2017action} to take previous actions into account. Eq.~\ref{eq: memory-dqn} defines the Q-value estimation with memory with the modified inputs.
\begin{align}\label{eq: memory-dqn}
    \Phi_t &= \phi(I_t, I_{t-1}, ..., I_{t-n}) \nonumber \\
    V(s) &= f_v(\Phi_t) \\
    A_{s,a} &= f_A(\Phi_t, a, (a_{t-1}, ..., a_{t-m})) \nonumber
\end{align}

The extracted features $\Phi_t$ from the current and previous frames are passed to the value estimator. $A(s,a)$ is defined by the action advantage estimator parameterized by the extracted features and previous actions.
The actions are fed to the model as concatenated one-hot-encoded vectors~\cite{murphy2012machine}. The setup is referred to as \textit{M-DQN}.

In order to address the sparsity of situations with valid stopping criteria the agent is exposed to (finding itself in a goal bin), we add a binary classifier to determine when the stopping criteria has been reached. By doing so, we modify the reward function detailed in Table~\ref{tab:reward_function} by removing the stopping decision. We call this \textit{MS-DQN}.

We train the feature extraction for all RL models and the binary classification network using a ResNet18 architecture~\cite{kaiming2015deep}.
Feature extraction is performed by removing the batch-normalization layers and the final average pooling layer to feed raw features into the state and advantage value estimators. 

\begin{figure*}[ht]
    \centering
    \framebox{
    \includegraphics[width=.24\textwidth]{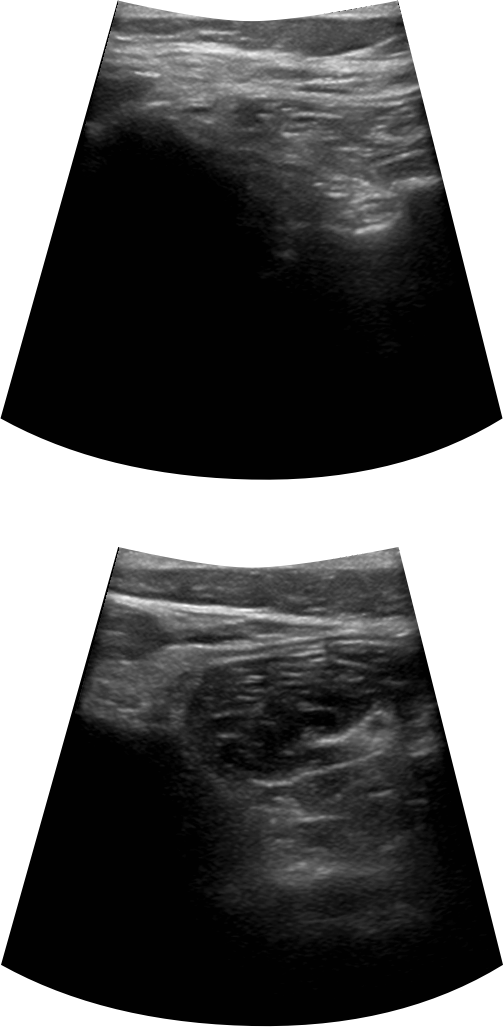}\hfill
    \includegraphics[width=.24\textwidth]{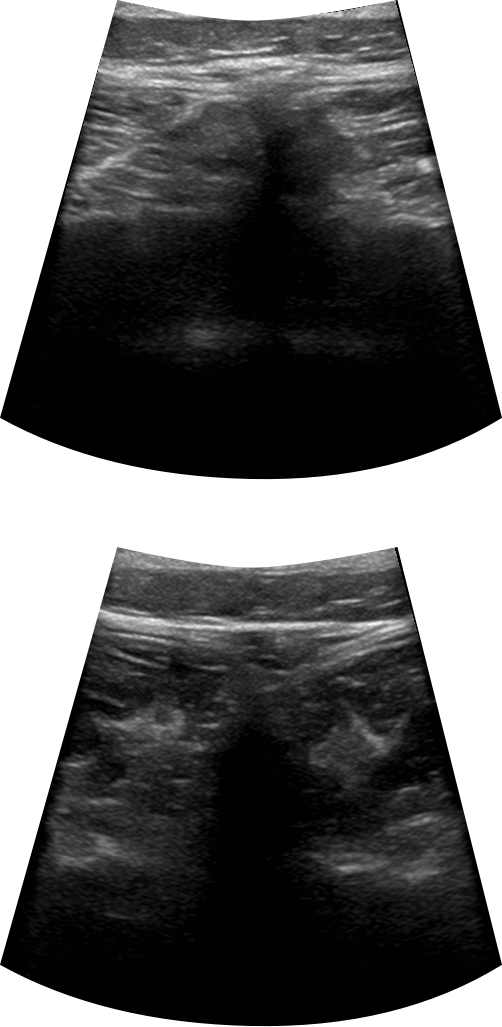}\hfill
    \includegraphics[width=.24\textwidth]{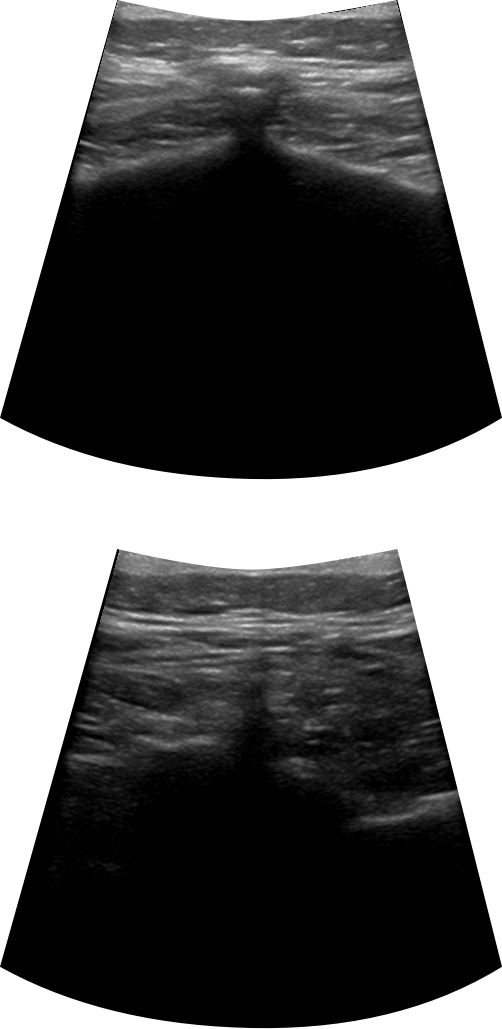}\hfill
    \includegraphics[width=.24\textwidth]{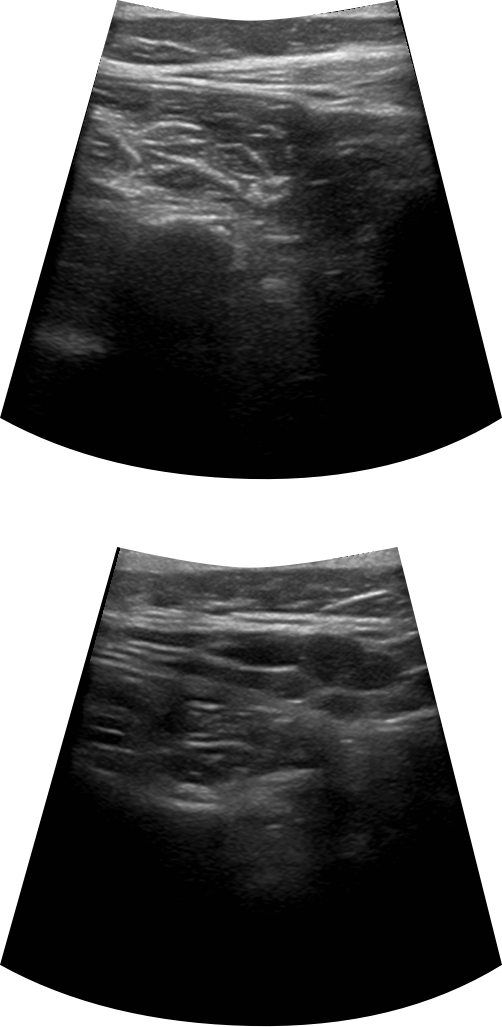}
    }
    \begin{minipage}[t]{.24\linewidth}
        \centering
        \subcaption{}\label{hip}
    \end{minipage}\hfill
    \begin{minipage}[t]{.24\linewidth}
        \centering
        \subcaption{}\label{spine}
    \end{minipage}\hfill
        \begin{minipage}[t]{.24\linewidth}
        \centering
        \subcaption{}\label{sacrum}
    \end{minipage}\hfill
        \begin{minipage}[t]{.24\linewidth}
        \centering
        \subcaption{}\label{soft}
    \end{minipage}
    \caption{The images above display exemplary US image samples from two of the subjects in the data-set. Each row belongs to one subject. The images correspond to (a) Left posterior pelvis; (b) L3 vertebra; (c) Sacrum; (d) Right lumbar region. In Fig.~\ref{fig:frame_grid}, the position of each frame in the projected grid on subjects is shown. These images show the variability of the same anatomical structure as seen in the US images between different subjects.}
    \label{fig:sample-frames}
\end{figure*}

\subsection{Problem setting}
With this work we aim at teaching an RL agent to successfully find the sacrum reacting only on information gained from US frames received, while navigating in the spinal region.
In other words, we aim to solve a search task with two degrees of freedom (DoF), on a defined plane situated parallel to the back of the subject. We call this plane, the \textit{parallel plane}.
To state our problem as an POMDP, we define the following terms:

\subsubsection{Action space}
The action space $\mathcal{S}_a$ is comprised of the actions ${up, down, left, right, stop}$ in the V-DQN and M-DQN.
In the case of MS-DQN, the $stop$ action is triggered by the binary classifier $f_{\text{stop}}()$.

\subsubsection{State}
The state of the environment is defined as the probe's position relative to the sacrum in the parallel plane. %on an x,y plane parallel to the back of the patient.
The state is fully defined by the position, thus complying with the Markovian property of the problem setting and the feasibility of using MDPs.

\subsubsection{Observation}
As our problem setting is modeled by a POMDP, the state is not known to the agent and needs to be estimated based on an observation $O(s)$ in the form of an US frame it receives from the environment.
The observations are defined by the state the agent finds itself in, while the observation defines the best action chosen by the agent.
Therefore, we can say that an agent that can estimate its state correctly is an agent that understands its environment and is more likely to successfully navigate towards its goal.
In our problem setting, the randomness in the observations comes from the anatomical differences and an eventual acquisition interference differences between subjects.

\subsubsection{Reward function}
We label bins that contain frames showing the sacrum as correct and defined numerical rewards given to the agent depending on direction of the actions in relation to the goals. The used reward function is detailed in table \ref{tab:reward_function}. The reward function heavily punishes incorrect stopping, as this would terminate the exploration in a wrong position. It also penalizes getting caught in back and forth movements, as by that behavior the agent would accumulate a net negative reward over time.

\begin{table}[thpb]
\caption{The reward function for the agent is defined by a discrete set of reward values. The values are defined as to heavily penalize incorrect stopping and strongly encourage correct stopping. The reward weights for the movement actions are selected so that inter-movement oscillatory motion is minimized.}
\label{tab:reward_function}
\begin{center}
\scalebox{1.3}{
\begin{tabular}{|c||c|}
\hline
Situation       & Reward   \\ 
\hline
Move closer     &  0.05     \\ 
\hline
Move away       & -0.1     \\ 
\hline
Correct stop    &  1.0      \\ 
\hline
Incorrect stop  & -0.25\\ 
\hline
\end{tabular}}
\end{center}
\end{table}

\subsubsection{Simulated robot navigation implementation}

We conduct simulated testing, by initializing our test environments at determined positions or states. We face our models with US frames obtained at that state and acted on the environment based on the action chosen by the agent. The simulated navigation is implemented as explained in Alg.~\ref{alg_navigation}.

\begin{algorithm}[h]
\SetAlgoLined
\KwResult{MS-DQN Robotic Navigation}
    $s_t = int(rand() * 164) $ \tcp*{init state}
    $t = 0$\;
    $t_{max} = 20$\;
    F = [] \tcp*{frame memory buffer}
    A = [] \tcp*{action memory buffer}\
    \While{$t < t_{max} \land a_t \in \mathcal{S}_a$}{
        $O_t = f_{E}(s_t) $ \tcp*{US frame}
        $a_t = f_{\text{stop}}(O_t)$ \tcp*{check stop}
        \If{$a_t \neq \textit{stop}$}{
            $a_t = argmax(f_{MS-DQN}(O_t))$ \tcp*{action}
            }
        \eIf{$a_t == \textit{stop}$}{
            break \tcp*{sacrum reached}
        }
        {
            $s_{t+1} = E(a_t)$ \tcp*{update state}
        }
        F[t] = $O_t$ \tcp*{frame to buffer}
        A[t] = $a_t$ \tcp*{action to buffer}
        $t = t + 1$
    }
\caption{Simulated Robot Navigation}
\label{alg_navigation}
\end{algorithm}

\section{Experimental setup}

\subsection{Project setup}

For data acquisition, we use a 7-axis robot certified for human interaction of the model  KUKA LBR iiwa 7 R800 manipulator (KUKA Roboter GmbH, Augsburg, Germany). The robot control runs on the Robotic Operating System (ROS)\footnote{\label{ros}http://www.ros.org/} using a custom software interface developed in our lab\footnote{\label{iiwa-stack}https://github.com/IFL-CAMP/iiwa\_stack}. The Ultrasound probe is attached to the end-effector with a 3D-printed mount.  To receive the US-frames, we used an Epiphan DVI2USB 3.0 frame-grabber (Epiphan Systems Inc. Palo Alto, California, USA) with a resolution of 800x600 pixels and a sampling frequency of 30 fps. We control the robot and process the images from a fixed workstation (Intel i5, NVIDIA GeForce GTX 1080). The image processing and robot control are implemented via custom software plugins integrated into the visualization framework ImFusionSuite\footnote{\label{imfusion}https://www.imfusion.de/} platform (ImFusion GmbH, Munich, Germany).

Ultrasound acquisitions are performed with a L8-3 linear US transducer and a Zonare z.one ultra sp Convertible Ultrasound System (ZONARE Medical Systems, Inc., Mountain View, California, United States). The imaging depth is set to 70 mm and an overall image gain of 90\%.
The robot is used with a compliant force control set to a maximum applied force of 2 N in the $z$ axis.

\subsection{Data-set} 

Our data-set\footnote{\label{data_set}https://github.com/hhase/sacrum\_data-set} collected in-house is comprised of US scans from the lower back of 34 volunteers in total. Each scan consists of eleven sweeps parallel to the spine with an off-set of 2 cm. We divide each sweep into 15 equally long segments and mapped the acquired frames to a grid of 11x15 bins. We fill each bin with five frames the agent would encounter when finding itself in that position. With this grid, we can simulate x-y navigation of the environment for training and testing the performance of the agent. In Fig.~\ref{fig:sample-frames}, we showcase different frames the agent could encounter in the grid.

We build one training set of 25 subjects containing a variety of acquisition qualities (artifacts, low resolution acquisitions, hard to recognize anatomies) to assure the model would be exposed to non-ideal training data. For validation and testing, we assemble a set of nine subjects with high quality scans (four and five respectively). We show the difference of the frames in Fig.~\ref{fig:sample-frames}.

\begin{figure}[thpb]
  \centering
  \framebox{
  \centering
  \includegraphics[scale=0.21]{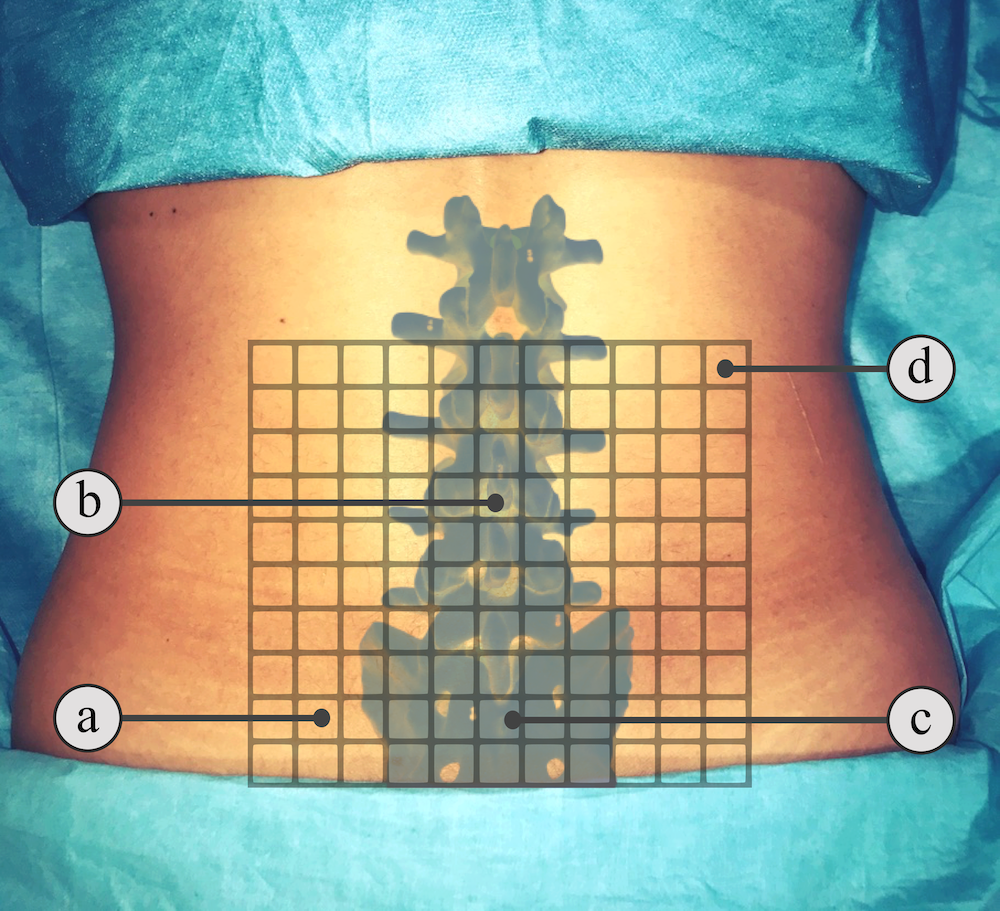}
  }
  \caption{Frame grid projected on the back of one of our volunteers. Here we show how the grid is positioned over the spine. The letters indicate where each sample frame in Fig.~\ref{fig:sample-frames} is approximately located.}
  \label{fig:frame_grid}
\end{figure}

\subsection{Implementation}

\subsubsection{Framework setup} Our framework is written on the deep learning (DL) library Tensorflow and extends RL-zoo~\cite{rl-zoo} and stable-baselines~\cite{stable-baselines}. Our code is publicly available on Github~\footnote{https://github.com/hhase/spinal-navigation-rl}.

\subsubsection{Model Training}
For training our models, we randomly initialize the agent in a random training environment and give it 50 attempts or \textit{steps} to reach the goal. We define this process as a \textit{training episode}. The training episode is terminated when either the agent chooses the \textit{stop} action or reaches the maximal permitted amount of steps. While training, the agent follows an $\epsilon$-greedy policy, meaning that the agent has a probability $\epsilon$ of behaving randomly, instead of choosing the action associated with the highest Q-value. By this, we address the exploration-exploitation dilemma~\cite{Sutton1998}, giving the agent a possibility to explore its environment to find eventual long term rewards. $\epsilon$ decays to 0.02 at a third of the total duration of the training.

For the binary classification model for stopping, we assign the frames containing a correct view of the sacrum to one class and the rest to another. For training, we over-sampled the underrepresented class (frames containing the sacrum) to compensate for the class imbalance. We augment the data-set with rotations and re-sized crops to generalize better. With this network, we obtain consistent accuracy of over $99\%$ on the test set.

Regarding the baseline, we use a standard DenseNet-121 architecture~\cite{huang2017densely} to train a classification network, where the predicted class corresponds to the chosen action. 
%In our problem setting, there are up to two possible correct movements at each position on the grid. To account for the multiple labels, we define the loss to be the minimum of the cross-entropy of the estimated output and each correct label.

\subsubsection{Metrics}
For testing our models, we initialize the agent in each of the 165 possible states of the unseen environments and give the agent 20 actions to reach the goal. We call each of these tests a \textit{run}.

As results, we report two performance indicators: policy correctness and reachability. 
To compute the policy correctness we define $n_c$ as the number of correct actions taken in the run $r$ and $n_t$ as the number of total actions taken in the run on environment $e$. 
$E$ is the total number of test environments and $R$ is the total amount of runs tried on each of them.
The policy correctness is computed as detailed in Eq.~\ref{policy_correctness}.

\begin{equation}
\label{policy_correctness}
\text{correctness} = \frac{1}{ER}\sum_{e=0}^{E}\sum_{r=0}^{R}\frac{n_c(e,r)}{n_t(e,r)}
\end{equation}

We define reachability as the ratio between runs that lead the agent to a stopping decision in a goal bin and the total number of runs. 
A run is not considered successful if the agent ends up in a goal bin but fails to stop.
To compute reachability we define $g$ as a boolean variable that is $1$ if the goal is reached in run $r$ on environment $e$ and $0$ if not. To compute the reachability we use Eq.~\ref{reachability}.

\begin{equation}
\label{reachability}
\text{reachability} = \frac{1}{ER}\sum_{e=0}^{E}\sum_{r=0}^{R}g(e,r)
\end{equation}

\begin{figure}[thpb]
  \centering
  \framebox{
  \centering
  \includegraphics[scale=0.09]{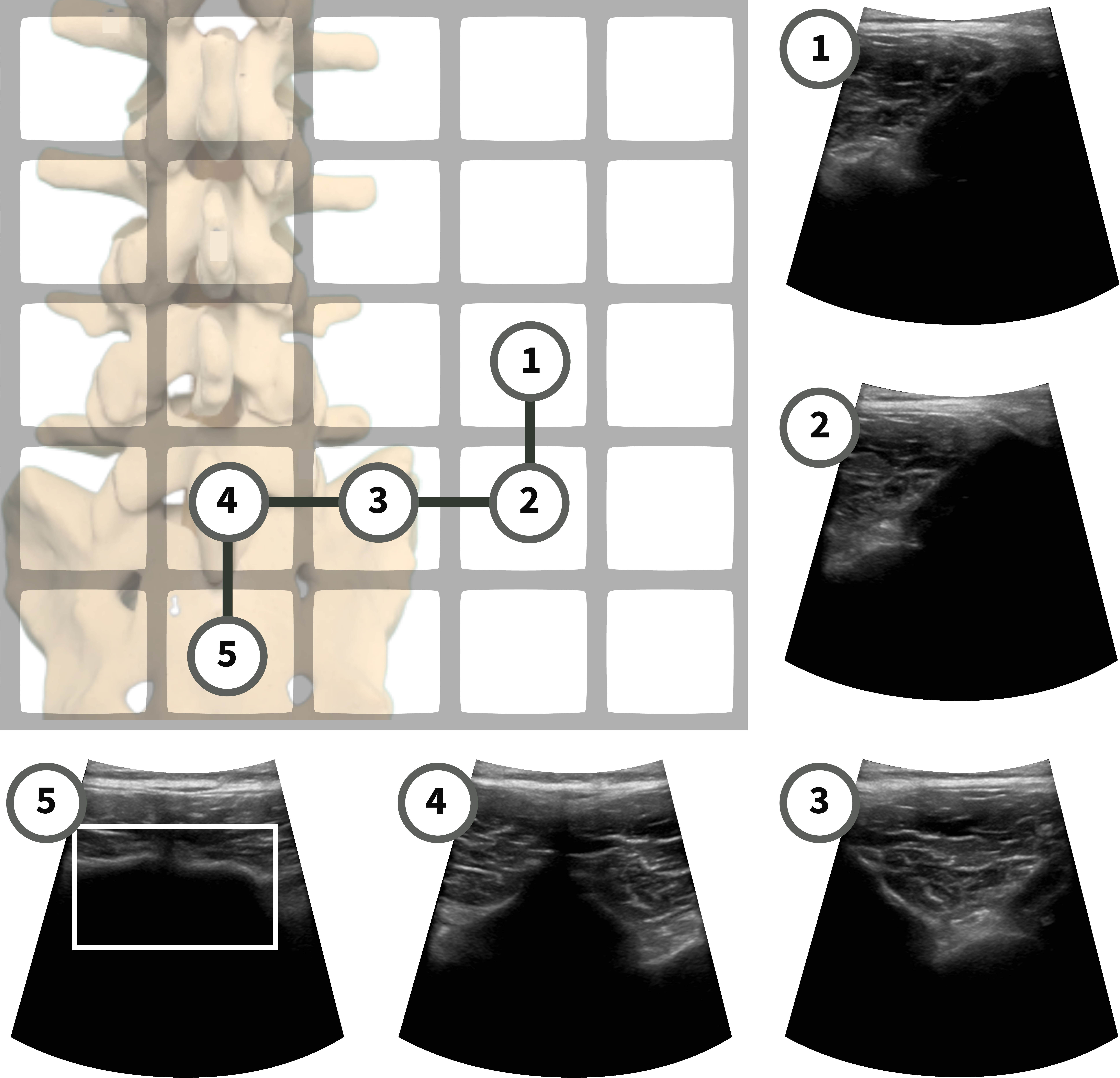}
  }
  \caption{Possible navigation sequence the agent would follow starting on the right lumbar region. The corresponding frames to the visited bins are labeled with the step number. In step number five the agent identifies a goal state. The sacrum is enclosed in the white bounding box.}
  \label{fig:navigation_path}
\end{figure}

In Fig. \ref{fig:navigation_path}, we show an example of a successfully testing run. In the case of this run, $n_c = n_t = 5$, as all the actions are taken in direction of the goal. Regarding reachability, $g(e,r) = 1$, because the agent successfully found the sacrum.

\section{Results and Discussion}
We choose the best model in each case, based on the median reachability value achieved on the validation set. We find that the median gives a more reliable measurement of the performance of the model, given the small validation set and strong subject dependency on the performance.

\begin{table}[h]
\caption{Performance of the different proposed architectures}
\label{result_table}
\begin{center}
\scalebox{1.2}{
\begin{tabular}{|c||c|c|}
\hline
NN architecture         & Policy correctness & Reachability\\
\hline
Classification CNN      &            58.42\% &      59.64\%\\
\hline
V-DQN                   &            55.37\% &      18.30\%\\
\hline
M-DQN                   &            49.49\% &      36.97\%\\
\hline
MS-DQN                  &   \textbf{79.53}\% & \textbf{82.91}\%\\
\hline
\end{tabular}}
\end{center}
\end{table}

To begin with discussing the results from table~\ref{result_table}, we can see that the V-DQN is outperformed by the M-DQN, by 20\% when it comes to reachability.
We attribute this to the inclusion of previous frames and actions. Now, the agent can recognize when it is stuck in a loop and break out of it. 
Therefore, the M-DQN can perform substantially better than the V-DQN in that aspect. 
However, the V-DQN still outperforms the M-DQN in terms of policy correctness by 6\%, and we can attribute this fact that the memory makes agent of the M-DQN follow sub-optimal paths when navigating towards the goal.
However, our proposed approach to combine a DQN with a memory buffer and a binary classifier for stopping, substantially outperforms the other baselines in both, policy correctness by 20 to 30\% and in reachability by 40 to 60\%.

These results signify the fact that the proposed RL approach is suitable for the task at hand since it delivers promising results in a challenging task like navigating the spinal region and successfully localizing the sacrum. 
We attribute the improvement to the inclusion of the binary classifier for stopping because, in our problem statement, the stopping action is the most difficult to achieve for pure DQL.
This difficulty arises because during the initial exploration phase during training, when following the $\epsilon$-greedy policy with a high probability of choosing random actions, the stopping action is most likely to be incorrect and thereby, heavily punished.
Also, because the reward function assigns comparatively large positive and negative rewards to the stopping action, the agent learns to avoid to stop when not entirely confident. 
The inclusion of a prioritized replay memory trying to counter the sparsity of transitions leading to a successful stop does not solve this shortcoming.

When looking at the classification network approach, we find that by not having memory, the classification agent easily gets stuck in loops and does not reach the goal. 
However, it proves to have better results when comparing to our V-DQN as its RL counterpart, as it is easier to train a classification network than a DQN.
The difference between SL and RL in visual navigation lays in the fact that SL decides the next-best-action based on features extracted from the input frame. In contrast, RL selects actions based on the estimated reward it can achieve from the state it is on.
Nonetheless, comparing our proposed DQN setup with the classification network, the results still highlight the advantage of RL for navigation tasks.

\begin{figure}[tpb]
    \centering
    \includegraphics[width=.24\textwidth]{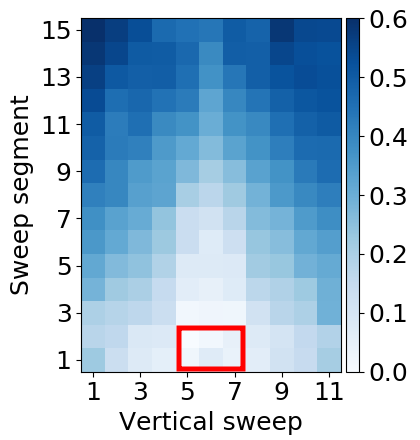}\hfill
    \includegraphics[width=.24\textwidth]{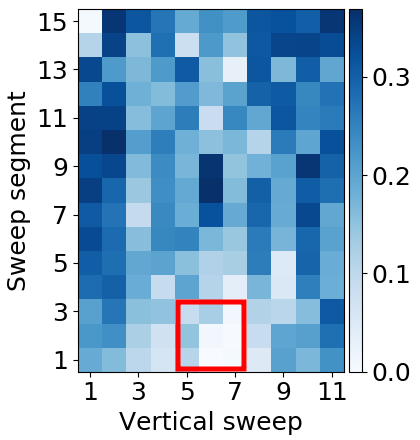}
    \newline
    \begin{minipage}[t]{.24\textwidth}
        \centering
        \subcaption{}\label{GT_state_val}
    \end{minipage}\hfill
    \begin{minipage}[t]{.24\textwidth}
        \centering
        \subcaption{}\label{V-DQN_state_val}
    \end{minipage}
    \newline
    \includegraphics[width=.24\textwidth]{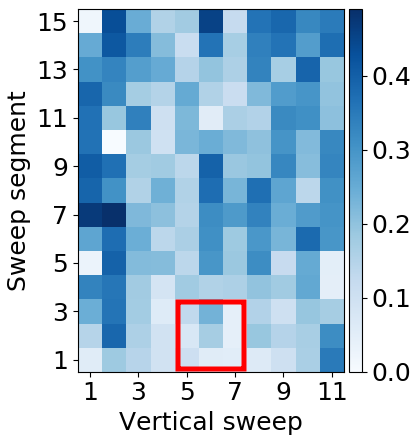}\hfill
    \includegraphics[width=.24\textwidth]{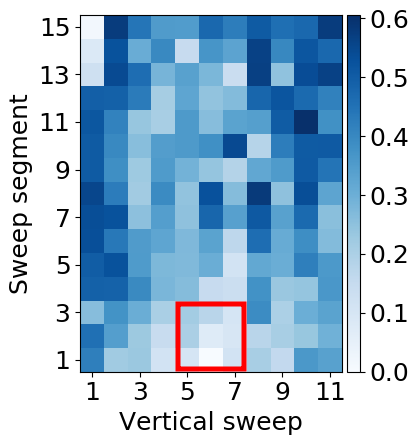}
    \newline
    \begin{minipage}[t]{.24\textwidth}
        \centering
        \subcaption{}\label{M-DQN_state_val}
    \end{minipage}\hfill
    \begin{minipage}[t]{.24\textwidth}
        \centering
        \subcaption{}\label{MS-DQN_state_val}
    \end{minipage}
    \caption{State value estimate maps. (a) corresponds to an training environment to showcase a ground truth to compare to the other state value maps obtained from the same test environment when using our three DQN setups. (b) is estimated with the V-DQN, (c) with the M-DQN and (d) with the MS-DQN. For this image, we subtracted the minimum state-value estimate of each map, to be able to compare them with the MS-DQN, as this setup does not have the rewards associated with stopping. The red bounding boxes show the goal bins.}
    \label{state-val-maps}
\end{figure}

A determining factor of the performance an RL agent has on unseen environments is the capability to correctly estimating the state it is in, as this gives the agent a notion on the value of its position within the environment. 
In Fig.~\ref{state-val-maps}, we show the state-value estimates on the same test environment for each of our DQN models. For comparison, we also show the state value estimates of one of our train environments as a ground truth. 
When comparing the ranges of the values on the different state value maps, we see that the only model achieving a similar range as the ground truth is our proposed MS-DQN.
The fact that the V-DQN is estimating worse than the M-DQN also reflects the results from table~\ref{result_table}. 

Besides the differences in the state-value estimations, we can see that it is hard to estimate state-values in unseen environments accurately. However, the ultimate goal of our models is mapping US-frames to actions. The information about the best action choice is contained in the advantage-value estimates, meaning that the agent is still able to take correct actions, despite being wrong about its state.

As shown in our results, however, pure RL struggles on its own with issues like reward sparsity and performance in unseen environments. Solving specific shortcomings of RL with SL proves to be very beneficial and needs to be explored further.

\section{CONCLUSIONS}

In this paper, we introduce a reinforcement learning-based ultrasound-guided robotic navigation. 
Despite the large anatomical variability within our volunteers, in a challenging task of spinal navigation to locate the sacrum, we showcased the superiority of our proposed approach against DQN and classification baselines. 
Introducing a binary classifier for deciding when to stop, brought substantial improvement to the method. 
Better results can be obtained by increasing our data-set. To move forward to an online implementation in a medical setting an ethical approval would be needed.

\printbibliography
\end{document}